\documentclass[10pt,twocolumn,letterpaper]{article}

\usepackage{times}
\usepackage{epsfig}
\usepackage{graphicx}
\usepackage{amsmath}
\usepackage{amssymb}
\usepackage{multirow}
\usepackage{array}
\usepackage{multirow}
\usepackage{lettrine}
\usepackage{flushend}
\usepackage{url}
\usepackage{colortbl}
\usepackage{color}
\usepackage{cancel}
\usepackage{fixltx2e}
\usepackage{tikz}
\usepackage{pgfplots}
\usepackage{pgfplotstable}
\usepgfplotslibrary{patchplots}

\usepackage[pagebackref=true,breaklinks=true,letterpaper=true,colorlinks,bookmarks=false]{hyperref}

\usepackage{cvpr}
\cvprfinalcopy % *** Uncomment this line for the final submission
\def\cvprPaperID{876} % *** Enter the CVPR Paper ID here

\ifcvprfinal\fi

\usepackage{pgfplots}
\usepackage{pgfplotstable}
\usepackage{pgf,tikz}

\usepgfplotslibrary{external}
\newcommand{\extfig}[2]{\tikzsetnextfilename{fig/extern/#1}{#2}}

\newcommand{\leg}[1]{\addlegendentry{#1}}

\begin{document}
\title{Asymmetric Feature Maps with Application to Sketch Based Retrieval}

\author{Giorgos Tolias \hspace{30pt} Ond{\v r}ej Chum\\
Visual Recognition Group, Faculty of Electrical Engineering, Czech Technical University in Prague\\
{\tt\small \{giorgos.tolias,chum\}@cmp.felk.cvut.cz}
}
\vspace{-10pt}

\maketitle

\newcommand{\real}{\mathbb{R}}
\newcommand{\integer}{\mathbb{Z}}

\def\sim{\mathcal{S}}

\def\Pc{\ensuremath{\mathcal{P}}\xspace}
\def\Qc{\ensuremath{\mathcal{Q}}\xspace}

\def\p{\ensuremath{p}\xspace}
\def\q{\ensuremath{q}\xspace}

\def\o{\ensuremath{\phi}\xspace}

\def\px{\ensuremath{\p_x}\xspace}
\def\py{\ensuremath{\p_y}\xspace}
\def\po{\ensuremath{\p_\o}\xspace}
\def\pw{\ensuremath{\p_w}\xspace}

\def\qx{\ensuremath{\q_x}\xspace}
\def\qy{\ensuremath{\q_y}\xspace}
\def\qo{\ensuremath{\q_\o}\xspace}
\def\qw{\ensuremath{\q_w}\xspace}

\newcommand{\mydelta}{{\mathchoice{\scriptstyle \Delta}{\scriptstyle \Delta}{\scriptscriptstyle \Delta}{\scriptscriptstyle \Delta}}}
\newcommand{\Dp}{\ensuremath{\mydelta \p}\xspace}
\newcommand{\Dq}{\ensuremath{\mydelta \q}\xspace}
\newcommand{\Dx}{\ensuremath{\mydelta x}\xspace}
\newcommand{\Dy}{\ensuremath{\mydelta y}\xspace}

\def\V{\mathbf{V}}

\def\l2{\ensuremath{\ell_2}\xspace}
\def\linf{\ensuremath{\ell_\infty}\xspace}

\renewcommand{\k}{\ensuremath{k}\xspace}
\newcommand{\ak}{\ensuremath{\hat{k}}\xspace}
\newcommand{\PSI}{\ensuremath{\V}\xspace}
\newcommand{\ith}{{(i)}}
\newcommand{\bOmega}{\ensuremath{\bar{\Omega}}\xspace}

\def\sssp{\hspace{1pt}}
\def\ssp{\hspace{3pt}}
\def\msp{\hspace{5pt}}
\def\bsp{\hspace{12pt}}

\def\etal{\emph{et al.}\xspace}
\def\ie{\emph{i.e.}\xspace}
\def\eg{\emph{e.g.}\xspace}

\newcommand{\alert}[1]{{\color{red}{#1}}}

\renewcommand{\paragraph}[1]{\vspace{.0\baselineskip}\noindent{\bf #1}\xspace}

\newcommand{\xcaption}[2][1]{\caption{#2}\vspace{-#1\baselineskip}}

\def\oversim#1#2{\lower .5pt\vbox{\lineskiplimit=\maxdimen \lineskip=.5pt
    \ialign{$\mathsurround=0pt #1\hfil ##\hfil $\crcr #2\crcr \rightarrow\crcr }}}
\def\plusarrow{\mathrel{\mathpalette\oversim{\!\scriptscriptstyle+}}}

\newcommand{\dpf}{\ensuremath{\bar{\sim}_{\scriptscriptstyle>}\!\! \plusarrow \! \bar{\sim}_{\scriptscriptstyle<}}}

\newcommand{\eq}{eqn.\xspace}

\begin{abstract}
We propose a novel concept of asymmetric feature maps (AFM), which allows to evaluate multiple kernels between a query and database entries without increasing the memory requirements. To demonstrate the advantages of the AFM method, we derive a short vector image representation that, due to asymmetric feature maps, supports efficient scale and translation invariant sketch-based image retrieval. Unlike most of the short-code based retrieval systems, the proposed method provides the query localization in the retrieved image. The efficiency of the search is boosted by approximating a 2D translation search via trigonometric polynomial of scores by 1D projections. The projections are a special case of AFM. An order of magnitude speed-up is achieved compared to traditional trigonometric polynomials. The results are boosted by an image-based average query expansion, exceeding significantly the state of the art on standard benchmarks.
\end{abstract}

\vspace{-1\baselineskip}
\vspace{-1ex}
\section{Introduction}
\label{sec:intro}
Efficient match kernel~\cite{BS09} is a popular choice in applications evaluating complex similarity measures on large collections of objects, where an object is a set of elements. This includes local feature descriptors~\cite{BS09,BTJ15} and image retrieval with short descriptors~\cite{TBFJ15}\footnote{The authors were supported by the MSMT LL1303 ERC-CZ grant.}.  

In efficient match kernel, all elements of the sets are mapped to a finite feature map~\cite{Rahimi-NIPS07,VZ12}. An inner product of the feature maps approximates evaluation of a specific kernel, defining similarity of the set elements.
We propose an extension to this concept. In the asymmetric feature map, the query uses a different embedding than the database objects. The query embedding defines the kernel that is evaluated between the query and the database entries. Thus, multiple kernels can be evaluated while the memory requirements for the database remains the same (up to a scalar per kernel) as for a single kernel to be evaluated.
The embeddings are obtained via joint kernel feature map optimization, which significantly improves the quality of kernel approximation for a fixed dimensionality of the feature map.

\begin{figure}[b]
\vspace{-2ex}
\begin{center}
\raisebox{1ex}{\includegraphics[height=0.10\columnwidth]{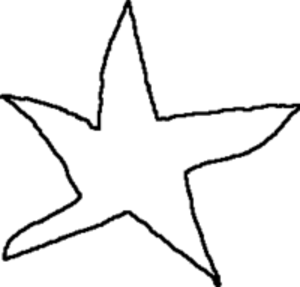}}\hspace{1ex}
\includegraphics[height=0.12\columnwidth]{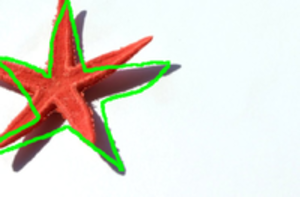}
\includegraphics[height=0.12\columnwidth]{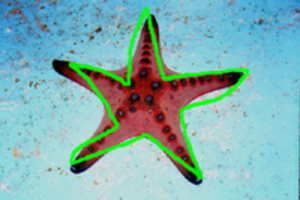}
\includegraphics[height=0.12\columnwidth]{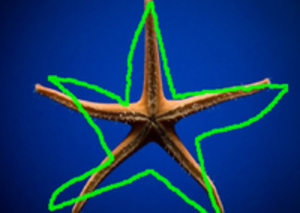}
\includegraphics[height=0.12\columnwidth]{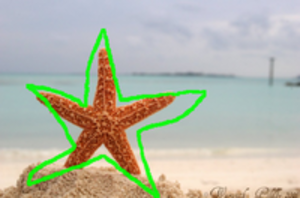}
\\ \vspace{0.2ex}

\raisebox{1ex}{\includegraphics[height=0.08\columnwidth]{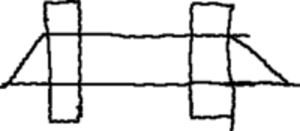}}\hspace{1ex}
\includegraphics[height=0.12\columnwidth]{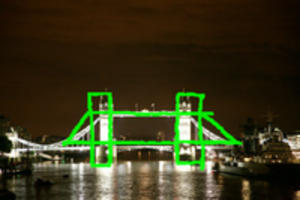}
\includegraphics[height=0.12\columnwidth]{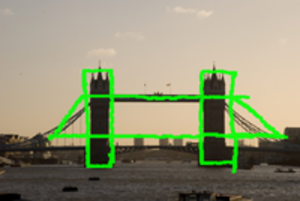}
\includegraphics[height=0.12\columnwidth]{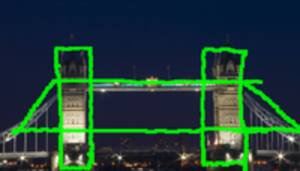}
\includegraphics[height=0.12\columnwidth]{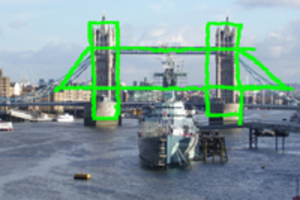}
\\ \vspace{0.2ex}

\end{center}
\vspace{-2.5ex}
\caption{Scale and translation invariant query-by-sketch retrieval. An example of sketch queries and top-retrieved images with the sketch localization overlaid in green color.
\label{fig:motiv}}
\end{figure}

The application domain of AFM is wide, in particular any method using efficient match kernel benefits from AFM. We evaluate the AFM on a sketch-based retrieval application.
Sketch-based retrieval has received less attention  than image retrieval and still remains challenging. Instead of a real image, the query consists of an abstract binary sketch. This allows the user to quickly outline an object, \eg by a finger on a tablet or smart phone, and search for relevant images (see Figure~\ref{fig:motiv}).
The progress in this area has more or less followed the footsteps of traditional image retrieval. The first systems employed global descriptors~\cite{CNM05}. Then, the Bag-of-Words paradigm with local descriptors and feature quantization~\cite{FFJS08,EHBA11,SBO15} was adopted. 

Due to the absence of textural cues on the query side, the image representations are shape based. Bridging the representation gap between hand-drawn sketches and real images is one of the challenges making the task difficult. Matching based on shape information has been addressed previously. For instance,  
in object recognition and detection~\cite{BBM05,FFJS08,KY11}, a costly online matching is performed, which prevented the methods to scale to large image collections.
Recent methods manage to index million~\cite{CWZZ11} to billion~\cite{SWXZ13} images for sketch-based retrieval, at the cost of sacrificed invariance to geometric transformations.

To demonstrate the impact of the AFM,
we propose a short vector image representation allowing to index large image collections for sketch-based search. Scale and translation invariant real-time search allows to process an order of millions of images per one processor thread.
The AFM based method achieves state-of-the-art results on standard benchmarks. The method runs at speed comparable to previously published approaches tailored to sketch-based search. Compared with methods based on efficient match kernel~\cite{TBFJ15}, the proposed method achieves order of magnitude speed-up.
Unlike most of the methods using low-dimensional descriptors, the proposed method delivers localization of the object in both scale and space.
The scale invariance is achieved by evaluating multiple kernels without the need to store multiple representations for database images.  
The translation invariance and object localization is provided by an efficient similarity evaluation on a 2D grid of translations. 
Namely, the four main contributions of this work are as follows.
(1) Asymmetric explicit feature maps allowing the use of multiple kernel functions without constructing multiple representations for database items are proposed.
(2) A joint kernel approximation approach for multiple kernels is derived, generalizing a recent approach of low dimensional explicit feature maps (LDFM)~\cite{Chum15}.
(3) The scoring through trigonometric polynomial introduced in~\cite{TBFJ15} 
is further extended and a significant speed-up of its evaluation is proposed. 
(4) State-of-the-art sketch-based image retrieval based on the AFM, which is further boosted by query 
expansion which acts, not on the edge maps as standard sketch matching, but on the original images.

The rest of paper is organized as follows. Related work is discussed in Section~\ref{sec:related} and the necessary background is presented in Section~\ref{sec:background}. Sections~\ref{sec:maps} and~\ref{sec:descriptor} describe our contributions on asymmetric explicit feature maps and on sketch retrieval, respectively, while the retrieval procedure and the experimental evaluation are analyzed in Section~\ref{sec:exp}.
\section{Related work}
\label{sec:related}

The most similar work to ours is the approach of Tolias~\etal~\cite{TBFJ15}, where the trigonometric polynomial scores were introduced in the context of image retrieval (see Section~\ref{sec:tripoly} for technical details). Shape properties of local features, such as dominant orientation or position, are jointly encoded with the SIFT descriptor. Despite initially assuming aligned objects, their kernel descriptor comes with an efficient way to compute similarity over multiple image transformations. Compared to their method, asymmetric feature maps introduced in our paper: i) reduce the memory requirements of multi-scale search by roughly a factor of 3, and ii) achieve an order of magnitude speed-up through approximate translation search.
The trigonometric polynomials have been also used by Bursuc \etal~\cite{BTJ15} in the context of rotation invariant feature descriptors.
The descriptor has recently shown competitive results with CNN based approaches~\cite{DescW16}.

Since we demonstrate the advantages of AFM on sketch based retrieval, we provide a brief review of relevant literature on this topic.
The line of research that focuses on sketches includes recognition~\cite{EHA12,YYSXH15} or retrieval~\cite{MYZR+13} of sketches. This paper addresses sketch-based image retrieval, which tries to match sketch queries to real images from a large collection. Following successful examples of traditional image retrieval, sketch-based methods employ global image representation~\cite{CNM05,Saaverda14} or local descriptors and the Bag-of-Words model. In the latter case,  representative methods employ local descriptors that are traditionally used on images~\cite{EHBA11,SBO15} or proposed particularly for this task~\cite{EHBA10,RDB11,RC13,CZL+13}. 
Some examples are HOG descriptors which are adapted for sketch retrieval~\cite{RC13} and were recently extended to capture color~\cite{BC15}, symmetry-aware and flip invariant descriptors~\cite{CZL+13}, and descriptors based on local contour fragments~\cite{RDB11}. Generic approaches performing learning of discriminative features have been shown effective for sketch retrieval too~\cite{SMGE11}.

Chamfer matching appears to be a good similarity measure for object shapes~\cite{TSTC03}. Recent attempts focus on Chamfer matching approximations in order to increase scalability. Cao \etal~\cite{CWZZ11} binarize the distance transform map and manage to index two million images. However, their approach completely lacks invariance. The same holds for the work of Sun \etal~\cite{SWXZ13} who increase the scale of the indexed collection up to one billion. Despite the achievement of scalability, rough approximations of Chamfer matching sacrifice accuracy. Recently, Parui and Mittal~\cite{PM14} proposed a similarity invariant approach able to index up to one million images. Their solution is based on dynamic programming to match chains of contour lines, while the main drawback is the costly off-line indexing. 

\section{Background} \label{sec:background}

We briefly review the necessary background, which includes efficient match kernels~\cite{BS09}, explicit feature maps \cite{VZ12} and efficient trigonometric polynomial scores~\cite{TBFJ15}.

\subsection{Efficient Match Kernels} \label{sec:EMK}

In many situations, an object is described by a set of measurements $\Pc = \{\p \in \real^d\}$. 
Employing a mapping $\Psi:~\real^d \rightarrow \real^D$ to the elements of $\Pc$, the set representation of efficient match kernels is defined as
\vspace{-5pt}
\begin{equation} \label{equ:emk}
\PSI(\Pc) =  \sum_{\p \in \Pc} \Psi(\p).
\vspace{-5pt}
\end{equation}
Then, a dot product between the set representation yields the similarity between sets
\vspace{-8pt}
\begin{equation}
\sim(\Pc,\Qc) = \PSI(\Pc)^\top \PSI(\Qc) = \sum_{\p \in \Pc} \sum_{\p \in \Qc} \Psi(\p)^\top \Psi(\q)\mbox{.}
\label{equ:setsim}
\vspace{-5pt}
\end{equation}
Normalized similarity is computed by cosine similarity~\cite{TBFJ15}, \ie, dot product of $\ell_2$ normalized vectors, 
\vspace{-8pt}
\begin{equation} \label{equ:cossim}
\bar{\sim}(\Pc,\Qc) = \frac{\PSI(\Pc)^\top \PSI(\Qc)}{\sqrt{\PSI(\Pc)^\top \PSI(\Pc)} \sqrt{\PSI(\Qc)^\top \PSI(\Qc)}},
\vspace{-5pt}
\end{equation}
while another choice is to normalize by the set cardinality~\cite{BS09}. Herein, the cosine similarity is adopted ensuring self-similarity is normalized to one. A number of image representations, such as BOW~\cite{SZ03,CDFWB04}, Fisher vectors~\cite{PD07}, or VLAD~\cite{JDSP10}, can be interpreted as efficient match kernels. 

\subsection{Explicit feature maps} \label{sec:featuremaps}

Let $K(\p, \q)$ be a one-dimensional ($p$ is now scalar) positive definite stationary kernel \cite{Scholkopf02} $K: \real \times \real \rightarrow \real$.
The value of a stationary kernel by definition depends only on the difference $\lambda = \p\!-\!\q$,
\vspace{-5pt}
\begin{equation}
% $
K(\p, \q) = K(\p, \p - \lambda) = \k(\lambda)\mbox{,}
% $
\vspace{-5pt}
\end{equation}
where $\k(\lambda)$ is a signature of kernel $K(\p, \q)$. Due to Bochner's theorem, kernel signature \k can be written as  
\vspace{-5pt}
\begin{equation}
  \k(\lambda) = \int_{0}^{\infty} \alpha(\omega) \cos(\omega \lambda) \, \mathrm{d}{\omega} \mbox{,}
\label{equ:kernel_spectrum}
\vspace{-5pt}
\end{equation}
where $\alpha(\omega) : \real^{+}_0 \rightarrow \real^{+}_0$.
The kernel signature is approximated by sum over a finite set $\Omega$ of frequencies  
\vspace{-5pt}
\begin{equation}
\ak(\lambda) \approx \sum_{\omega \in \Omega} \alpha_\omega \cos(\omega \lambda) \mbox{,}
\label{equ:kernel_approx}
\vspace{-5pt}
\end{equation}
where $\alpha_\omega \in \real^{+}_0$.
Applying the trigonometric identity 
\vspace{-5pt}
\begin{equation}
% $
\cos(\p-\q) = \cos(\p)\cos(\q) + \sin(\p)\sin(\q)
% $
\vspace{-5pt}
\end{equation}
gives rise to feature map (or feature embedding) $\Psi_\omega: \real \rightarrow \real ^ 2$ defined as
\vspace{-5pt}
\begin{equation}
\Psi_\omega(\p) = \big(\sqrt{\alpha_\omega}\cos(\omega \p), ~\sqrt{\alpha_\omega}\sin(\omega \p)\big) ^\top. 
\label{equ:map_sym}
\vspace{-5pt}
\end{equation}
The inner product of two such vectors reconstructs the terms of equation (\ref{equ:kernel_approx}) since
$\Psi_\omega(\p)^\top \Psi_\omega(\q) = \alpha_\omega \cos(\omega (\p-\q))$. Let the feature map $\Psi(\p): \real \rightarrow \real^D$ be constructed as a concatenation of $\Psi_\omega(\p)$ for all $\omega \in \Omega$.
Now, the inner product
\vspace{-10pt}
\begin{equation}
  \Psi(\p)^\top \Psi(\q) = \ak(\p-\q) \approx K(\p, \q)
\vspace{-5pt}
\end{equation}
 evaluates the approximation of the kernel signature (\ref{equ:kernel_approx}) and hence approximates the original kernel  $K$.
 The choice of the number of frequencies $|\Omega|$ determines the quality of the approximation and  the dimensionality of the embedding.
The dimensionality is $2 |\Omega|$, or $2 |\Omega|-1$ if $0 \in \Omega$\footnote{If $0 \in \Omega$, then $\alpha_0 \sin(0 \lambda) = 0$ for all $\lambda$ can be dropped from the explicit feature map.}.

\paragraph{Feature map construction.} We mention in detail (and compare) two approaches to construct the explicit feature maps. We do not consider random feature maps~\cite{Rahimi-NIPS07}, which approximate the integral in (\ref{equ:kernel_spectrum}) using Monte-Carlo methods.
Such feature maps provide a poor approximation for low-dimensional feature maps. 

Vedaldi and Zisserman~\cite{VZ12} propose the following approximation to a kernel signature $\k(\lambda)$ on an interval $\lambda \in [- \Lambda, \Lambda]$. First, a periodic function $g$ with period $2\Lambda$ is constructed, so that $g(\lambda) = k(\lambda)$ for $\lambda \in [ -\Lambda, \Lambda ]$. The feature map is then efficiently obtained by approximating periodic $g$ using harmonic frequencies only. This approach has been shown sub-optimal \cite{Chum15}. Further, the periodic function $g$ is not even guaranteed to be positive definite.  

A convex optimization approach is proposed by Chum~\cite{Chum15}. The input domain of $\ak(\lambda)$ is discretized to finite set $Z \subset [ 0, \Lambda ]$.
The quality of the approximation is measured at points in $Z$ as, for example, an \linf norm
\vspace{-5pt}
\begin{equation}
C_\infty(\k, \ak)  =  \max_{\lambda \in Z}\  |\k(\lambda) - \ak(\lambda)| \mbox{.}  \label{eqn:inferr}
\vspace{-5pt}
\end{equation}
The set of frequencies $\Omega \subset \bOmega$ are selected from a pool of frequencies $\bOmega$, and corresponding weights $\alpha_\omega \ge 0$, $\omega \in \bOmega$ jointly through a solution of a linear program  
\begin{equation}
\min_{\k}  C(k,\ak) + \gamma \mbox{$\sum_{\omega \in \bOmega}$} \ \alpha_\omega \mbox{,} 
% $$
\vspace{-5pt}
\end{equation}
where $\gamma \in \real^+$ is a weight on the $l_1$ regularizer controlling the trade-off between the quality of the approximation and the sparsity of $\alpha_\omega$.
This is the method we adopt and extend in this work.
 
\subsection{Alignment using trigonometric polynomials} \label{sec:tripoly}
Tolias \etal~\cite{TBFJ15} propose an image representation derived by efficient match kernels and explicit feature maps. We focus on the case that all measurements of set $\Pc$ are shifted by a constant value \Dp; note that measurements $p$ are now scalars. The similarity under such shift forms a trigonometric polynomial 
\vspace{-5pt}
\begin{equation}
\sim(\Pc_{\Dp}, \Qc) = \sum_{\omega \in \Omega} \left( \beta_\omega \cos(\omega \Dp)  +  \gamma_\omega \sin(\omega \Dp) \right),
\vspace{-5pt}
\end{equation}
with $\Pc_{\Dp} = \{\p-\Dp, \p \in \Pc \}$. Parameters $\beta_\omega$ and $\gamma_\omega$ are given by dot products of relevant sub-vectors of $\V(\Pc)$ and $\V(\Qc)$. Finally the similarity measure that is invariant under such shifting is given by $\sim_1(\Pc_{\Dp}, \Qc) = \max_{\Dp} \sim(\Pc_{\Dp}, \Qc)$.

We postpone further analysis of polynomials of scores until the image representation is introduced in Section~\ref{sec:descriptor}.

% ------------------------------------------------
\section{Asymmetric feature maps}
\label{sec:maps}

In this section, we introduce the concept of asymmetric feature maps. Unlike in classical explicit feature maps, a different feature map $\hat{\Psi}$ is used on the query side and a different one $\hat{\Psi}'$ is used on the database side. We show that with asymmetric feature maps, a number of different kernels can be efficiently evaluated between query and database vectors while keeping the database storage of fixed size.
Compare the feature map in equation (\ref{equ:map_sym}) to the following feature maps for the query and database side respectively
\vspace{-14pt}
\begin{align}
\hat{\Psi}_\omega(\q) & = \big(\alpha_\omega\cos(\omega \q), ~\alpha_\omega\sin(\omega \q)\big) ^\top \label{equ:map_asym1}\\
\hat{\Psi}^{\prime}_\omega(\p)& = \big(\cos(\omega \p), ~\sin(\omega \p)\big) ^\top \mbox{.}
\label{equ:map_asym2}
\vspace{-14pt}
\end{align}
The inner products $\hat{\Psi}(\q)^{\top} \hat{\Psi}^{\prime}(\p) = \Psi(\q)^{\top} \Psi(\p)$ are preserved. The kernel function is fully defined by the weights on the query side. No additional storage is required on the database side to evaluate the kernel. The same holds for efficient match kernels, as (\ref{equ:emk}) is a normalized sum of feature maps. To evaluate the cosine similarity (\ref{equ:cossim}),
 only a single scalar per kernel $K^\ith$ needs to be stored for each database entry $\Pc$ -- the $\ell_2$ norm $\sqrt{\PSI^\ith(\Pc)^\top \PSI^\ith(\Pc)}$, which is computed offline.

\paragraph{Joint approximation of multiple kernels.}
In order to evaluate a number of different kernels $K^\ith(\p, \q)$ using the asymmetric feature maps, all respective explicit feature maps $\Psi^{\ith}$ have to be based on the same set of frequencies $\Omega$. 
A naive approach would be to optimize the set of frequencies for one of the kernels and keep it fixed for other kernels. This approach, however, leads to poor approximation, as shown in Figure~\ref{fig:jointapprox}.
We propose an extension to LDFM~\cite{Chum15} to jointly approximate a set of kernels $K^{\ith}$ represented by their respective kernel signatures $\k^\ith$, $i \in \{1 \ldots n\}$.
The quality of the approximation is measured by the sum of individual qualities (\ref{eqn:inferr}) 
\vspace{-5pt}
\begin{equation*}
C^*_\infty = \sum_{i=1}^n C_\infty(\k^\ith, \ak^\ith) \! = \! \sum_{i=1}^n \max_{\lambda \in Z}\  |\k^\ith(\lambda) - \ak^\ith(\lambda)| \mbox{.}  \label{eqn:jointerr}
\vspace{-8pt}
\end{equation*}
The optimization is performed by executing a linear program \\[-1.2\baselineskip]
\begin{equation}
\min_{\alpha^\ith_\omega |\ \omega \in \Omega}  C^*_\infty + \gamma \sum_{\omega \in \Omega} \max_i \alpha^\ith_\omega \mbox{,} 
\end{equation}
where $\gamma$ is a weight of the sparsity regularizer that controls the number of frequencies used, \ie the dimensionality of the feature map.
Following the approach of Chum~\cite{Chum15}, to ensure the required dimensionality of the feature map, a binary search for $\gamma$ is performed. 

Figure~\ref{fig:jointapprox} presents the approximation of three different kernels using the same set of frequencies. We compare the approximation using only harmonic frequencies, the naive approximation mentioned above, and our joint approximation. The latter has a significantly better fit.

\begin{figure*}[t]
\vspace{-2.5ex}
\centering
\pgfplotsset{                            %%% newly added for font family setting
      height=0.22\textwidth,
      width=0.33\textwidth,
		xlabel={$\lambda$},
      legend pos=north east,
      legend cell align=left,
      legend style={font =\scriptsize, fill opacity=0.8, row sep=-3pt},
      grid=both,
      ymin = -0.28, ymax = 1.15,
      ytick={0, 0.5, ..., 1}, 
   	xmin = -0.05, xmax = 3.2,      
      xtick={0,    0.7854,    1.5708,    2.3562,   3.1416},
      xticklabels={$0$, $\frac{\pi}{4}$, $\frac{\pi}{2}$, $\frac{3\pi}{4}$, $\pi$},
      xlabel style={yshift=1ex,xshift=1.5ex},      
      ylabel style={yshift=-2.5ex},      
      title style={yshift=-1.5ex},      
 }
% \tikzstyle{every node}=[font=\large]
\extfig{joint_opt_small}{
\begin{tikzpicture}
   \begin{axis}[%
	   title={$\sigma=0.12$},
      ylabel={$k(\lambda)$}]
\addplot[color=blue,line width=1pt] table[x index=0,y index=1]{pgf/data/joint_opt.dat};\leg{$k(\lambda)$}
\addplot[color=orange,line width=1pt] table[x index=0,y index=13]{pgf/data/joint_opt.dat};\leg{$\hat{k}(\lambda)$, Fourier}
\addplot[color=green,line width=1pt] table[x index=0,y index=4]{pgf/data/joint_opt.dat};\leg{$\hat{k}(\lambda)$, naive}
\addplot[color=red,line width=1pt] table[x index=0,y index=7]{pgf/data/joint_opt.dat};\leg{$\hat{k}(\lambda)$, joint}
\end{axis}
\end{tikzpicture}
}
\extfig{joint_opt_med}{
\begin{tikzpicture}
   \begin{axis}[%
   	  title={$\sigma=0.16$}]
\addplot[color=blue,line width=1pt] table[x index=0,y index=2]{pgf/data/joint_opt.dat};%\leg{$k(\lambda)$}
\addplot[color=orange,line width=1pt] table[x index=0,y index=14]{pgf/data/joint_opt.dat};%\leg{$\hat{k}(\lambda)$, Fourier}
\addplot[color=green,line width=1pt] table[x index=0,y index=5]{pgf/data/joint_opt.dat};%\leg{$\hat{k}(\lambda)$, naive}
\addplot[color=red,line width=1pt] table[x index=0,y index=8]{pgf/data/joint_opt.dat};%\leg{$\hat{k}(\lambda)$, joint}
\end{axis}
\end{tikzpicture}
}
\extfig{joint_opt_large}{
\begin{tikzpicture}
   \begin{axis}[%
   	  title={$\sigma=0.20$}]
\addplot[color=blue,line width=1pt] table[x index=0,y index=3]{pgf/data/joint_opt.dat};%\leg{$k(\lambda)$}
\addplot[color=orange,line width=1pt] table[x index=0,y index=15]{pgf/data/joint_opt.dat};%\leg{$\hat{k}(\lambda)$, Fourier}
\addplot[color=green,line width=1pt] table[x index=0,y index=6]{pgf/data/joint_opt.dat};%\leg{$\hat{k}(\lambda)$, naive}
\addplot[color=red,line width=1pt] table[x index=0,y index=9]{pgf/data/joint_opt.dat};%\leg{$\hat{k}(\lambda)$, joint}
\end{axis}
\end{tikzpicture}
}
\vspace{-1.5ex}
\xcaption{Approximation comparison of multiple 1D RBF kernels (with different $\sigma$) using the same set of frequencies. We show approximation using harmonic frequencies only, a naive approach of optimizing the leftmost kernel and using the same frequencies for all, and our joint approximation. Maximum value is normalized to one such that the errors are comparable. $|\Omega|=7$ for all approximations.} \label{fig:jointapprox}
\end{figure*}
\section{Sketch-Based Retrieval}
\label{sec:descriptor}
In this section we present our sketch descriptor employing explicit feature maps and elaborate on the efficient trigonometric polynomial of scores to further approximate it. Our methodology is presented for the symmetric feature maps, while the asymmetric case is equivalent.
We finally present efficient ways to perform the initial ranking and re-ranking for sketch-based image retrieval.

\subsection{Sketch descriptor}
Consider a binary sketch as a set of contour points, that is a set of pixels $\Pc$ that lie on the contour. A \emph{contour pixel} $\p \in \Pc$ is represented as $\p = (\px,\py,\po,\pw)$, where \px and \py are 2D image coordinates, \po is the gradient angle (or orientation) of the contour at $(\px, \py)$, and \pw is a strength of the gradient. For real images, the contour parameters are obtained form an edge detector. For sketches, $\pw = 1$ is set for all contour pixels.

The similarity between contour pixels is computed using a multiplicative kernel composed of three one-dimensional kernels, spatial kernels over \px, \py, and an orientation kernel over \po. The 1D stationary kernels are denoted $K_x(\px, \qx) = \k_x(\lambda_x)$, $K_y(\py, \qy) = \k_y(\lambda_y)$, and $K_\o(\po, \qo) = \k_\o(\lambda_\o)$ respectively. 
The \emph{sketch descriptor} is a weighted sum of contour pixel feature maps\footnote{We use $\Psi$ to denote both the spatial and orientation feature map and simplify the notation. In fact, $\Psi(\px)$ and $\Psi(\py)$ approximate the spatial kernels $\k_x$ and $\k_y$, respectively, which are identical, while $\Psi(\po)$ the orientation kernel $\k_\o$.}
\vspace{-5pt}
\begin{equation}
\V(\Pc) = \sum_{\p \in \Pc} \pw \Psi(\px) \otimes \Psi(\py) \otimes \Psi(\po). \label{eqn:imgrepre}
\vspace{-5pt}
\end{equation}
It is easy to show that \emph{sketch similarity} (\ref{equ:setsim}) becomes
\vspace{-5pt}
\begin{equation}
\sim(\Pc,\Qc) = \sum_{\p \in \Pc} \sum_{\q \in \Qc} \pw \qw \k_x(\lambda_x) \k_y(\lambda_y) \k_\o(\lambda_\o).
\vspace{-5pt}
\end{equation}

The orientation and spatial kernels are implemented by 1D RBF kernels with parameters $\sigma_\o$ and $\sigma_x=\sigma_y$, respectively. The set of frequencies are denoted by $\Omega_\o$ and $\Omega_x=\Omega_y$, while the dimensionality of the corresponding embeddings is $D_x=2|\Omega_x|-1$ and $D_\o=2|\Omega_\o|-1$, respectively. Note that frequency $\omega=0$ is always included. The sketch descriptor has dimensionality ${D_x}^2 D_\o$. 

The proposed representation constitutes a holistic representation encoding the global sketch shape. We now define a representation encoding only one of the spatial coordinates along with the orientation. It is equivalent to the projection of contour pixels on the horizontal/vertical image axis. The sketch descriptor derived by projection on the horizontal axis is given by 
\vspace{-5pt}
\begin{equation}
\V_x(\Pc) = \sum_{\p \in \Pc} \pw \Psi(\px) \otimes 1 \otimes \Psi(\po) \mbox{,}
\vspace{-5pt}
\end{equation}
where the $\otimes 1$ can be omitted and is only used to show, that the $x$-projection is a sub-vector of (\ref{eqn:imgrepre}) and hence a special case of the proposed asymmetric feature map. This stems from the presence of the constant component of the feature map for $y$, corresponding to $0 \in \Omega_y$. An analogous derivation holds for $\V_y(\Pc)$ and vertical projection. 

\subsection{Position alignment}
The sketch descriptor encodes spatial coordinates and orientation of contour pixels. Therefore, alignment of objects is assumed, \ie centered and up-right objects. Such an assumption does not hold in real image collections and introduces significant limitations. We now detail the  polynomial of scores (mentioned in Section~\ref{sec:background}) proposed by Tolias \etal~\cite{TBFJ15}. We show that translation invariance is achieved by polynomial of scores, and that its evaluation can be efficiently approximated to speed up the search process.

\paragraph{One dimensional.}
Consider the $x$-projected sketch descriptor $\V_x(\Pc)$. Let $\Pc_{\Dx}$ be the shifted version sketch $\Pc$ where all contour pixels are horizontally translated by $\Dx$. Elementary trigonometric identities allow us to show that
\vspace{-3pt}
\begin{align}
\Psi_{\omega}^c(x\!-\!\Dx)\! &= \Psi_{\omega}^c(x) \cos(\omega \Dx) + \Psi_{\omega}^s(x) \sin(\omega \Dx) \nonumber \\ 
\Psi_{\omega}^s(x\!-\!\Dx)\! &=  \Psi_{\omega}^s(x) \cos(\omega \Dx) - \Psi_{\omega}^c(x) \sin(\omega \Dx),
\vspace{-3pt}
\end{align}
where $\Psi_{\omega}^c$ and $\Psi_{\omega}^s$ denote the first and second dimension of $\Psi_{\omega}$~(\ref{equ:map_sym}), respectively. Let $\V_\omega^c(\Pc)$ be the sub-vector of $\V(\Pc)$ comprised all elements that contain term $\Psi_\omega^c(x)$, and similarly for $\V_\omega^s(\Pc)$. It turns out that the descriptor of the translated sketch is constructed from that of the original sketch
\vspace{-3pt}
\begin{align}
\V_\omega^c(\Pc_{\Dx})\!=& \V_\omega^c(\Pc) \cos(\omega \Dx) + \V_\omega^s(\Pc) \sin(\omega \Dx) \nonumber \\ 
\V_\omega^s(\Pc_{\Dx})\!=& \V_\omega^s(\Pc) \cos(\omega \Dx) - \V_\omega^c(\Pc) \sin(\omega \Dx).
\label{equ:vectrans}
\vspace{-3pt}
\end{align}
The sketch similarity between sketches $\Pc$ and $\Qc$ under horizontal translation $\Dx$ is a trigonometric polynomial\\[-.5\baselineskip]  
\vspace{-3pt}
\begin{equation}
\sim(\Pc_{\Dx}, \Qc) = \sum_{\omega \in \Omega_x} \left( \beta_\omega \cos(\omega \Dx)  +  \gamma_\omega \sin(\omega \Dx) \right) \mbox{,}
\label{equ:tripol1sm}
% \vspace{-.3\baselineskip}
\vspace{-3pt}
\end{equation}
 with coefficients  $\beta_\omega$ and $\gamma_\omega$ 
\vspace{-3pt}
\begin{align}
\beta_\omega =& \V_\omega^c(\Pc)^\top \V_\omega^c(\Qc) + \V_\omega^s(\Pc)^\top \V_\omega^s(\Qc) \nonumber \\ 
\gamma_\omega =& \V_\omega^s(\Pc)^\top \V_\omega^c(\Qc) - \V_\omega^c(\Pc)^\top \V_\omega^s(\Qc).
\label{equ:tripol1}  
\vspace{-3pt}
\end{align}
The coefficients $\beta_\omega$ and $\gamma_\omega$ of this polynomial are computed by two products of sub-vectors with $D_\o$ dimensions. In total there are $N_1=D_x$ coefficients to be computed. Finally, similarity for any translation  with (\ref{equ:tripol1sm}) has cost equal to $N_1$ scalar multiplications. If the candidate translations are fixed, then terms $\cos(\omega \Dx)$ and $\sin(\omega \Dx)$ can be pre-computed. Normalized similarity comes at no extra cost since the \l2 norm of sketch descriptor remains constant under translations ($\k_x$ is a stationary kernel):
\vspace{-5pt}
\begin{equation}
\V(\Pc_{\Dx})^\top \V(\Pc_{\Dx}) = \V(\Pc)^\top \V(\Pc).
\vspace{-4pt}
\end{equation}
Similarity that is invariant to horizontal translation is computed by maximizing (\ref{equ:tripol1sm}) for all possible translations 
\vspace{-5pt}
\begin{equation}
\sim_{x}(\Pc_{\Dx}, \Qc) = \max_{\Dx} \sim(\Pc_{\Dx}, \Qc).
\vspace{-4pt}
\end{equation}
Note that this similarity is also invariant to vertical translation as $y$ coordinate is not encoded at all. However, this makes the representation less discriminative. The actual \emph{sketch transformation} aligning the two shapes is given by $\hat{x}_1=\arg\max_{\Dx} \sim(\Pc_{\Dx}, \Qc)$. Similarity based on the vertical projection is defined in a similar way. 
\begin{figure}[t]
\setlength{\fboxsep}{10pt} \setlength{\fboxrule}{1pt}
\raisebox{3ex}{\fbox{\includegraphics[height=0.11\columnwidth]{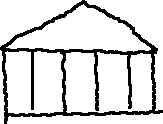}}}\hfil
\setlength{\fboxsep}{0pt} \setlength{\fboxrule}{1pt}
\fbox{\includegraphics[height=0.23\columnwidth]{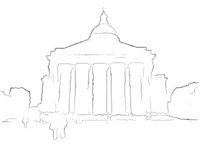}}\hfil
\includegraphics[height=0.23\columnwidth]{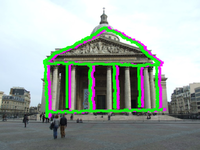}
\caption{Sketch (left) and the edge map (middle) of a real image (right). 
We depict the translations maximizing similarity based on 1D projections (magenta) and the full 2D case (green). 
\label{fig:sketchmatch}
\vspace{-3ex}}
\end{figure}
\begin{figure*}[t]
\vspace{-2.5ex}
\begin{center}
\begin{tabular}{@{\sssp}c@{\sssp}c@{\sssp}c@{\sssp}}
\extfig{align_example_sim1D}{
\pgfplotsset{                            %%% newly added for font family setting
      label style = {font=\footnotesize},
      tick label style  = {font=\footnotesize},
      every axis label/.append style = {font=\footnotesize}
}
\begin{tikzpicture}
% \tikzstyle{every node}=[font=\Huge]
   \begin{axis}[%
      height=0.22\textwidth,
      width=0.32\textwidth,
      xlabel={\color{red}{$\Delta y$}~~~\color{blue}{$\Delta x$}},
      ylabel={\color{red}{$\bar{\mathcal{S}}(\mathcal{P}_{\Delta y}, \mathcal{Q})$}}~~~~\color{blue}{$\bar{\mathcal{S}}({P}_{\Delta x}, \mathcal{Q})$},
      legend pos=south east,
      legend cell align=left,
      legend style={font=\footnotesize, fill opacity=0.7},
      grid=both,
      xmin = -120, xmax = 120,
      xtick={-120, -60, ..., 120},
	  ymin = -0.0, ymax = 1,	  
	  ytick={0, 0.2, ..., 1},
 	  ytick align=inside,
	  xtick align=outside,
     ylabel style={yshift=-2.5ex},      
     xlabel style={yshift=1ex},      
   ]
\addplot[color=blue,smooth, line width=2pt] table[x index=0,y index=1]
{pgf/data/38_4021_1dx.dat};\leg{horizontal}
\addplot[color=red,smooth, line width=2pt] table[x index=0,y index=1]
{pgf/data/38_4021_1dy.dat};\leg{vertical}
\end{axis}
\end{tikzpicture}
} &
\extfig{align_example_sim2D}{
\pgfplotsset{                            %%% newly added for font family setting
      label style = {font=\footnotesize},
      tick label style  = {font=\footnotesize},
      every axis label/.append style = {font=\footnotesize}
}

\begin{tikzpicture}
%\tikzstyle{every node}=[font=\Large]
   \begin{axis}[%
      view={30}{40},
      height=0.25\textwidth,
      width=0.32\textwidth,
      xlabel={$\Delta y$},
      ylabel={$\Delta x$},
      zlabel={$\bar{\mathcal{S}}(\mathcal{P}_{\Delta x, \Delta y}, \mathcal{Q})$},
      xmin = -120, xmax = 120,
      xtick={-100, -50, ..., 100},
      ymin = -120, ymax = 120,
      ytick={-100, -50, ..., 100},
 	  ztick=\empty,
	  xtick align=inside,
	  ytick align=outside,
	  zlabel shift = 2ex,
  	  xlabel shift = -3ex,
  	  ylabel shift = -3ex,
      zlabel style={yshift=-2ex,xshift=2ex}      
   ]
\addplot3 [surf, mesh/rows=45, patch type=bilinear] table[x index=0, y index=1, z index=2] {pgf/data/38_4021_2d.dat};
 \end{axis}
\end{tikzpicture}
} &
\extfig{align_example_sim2D_binary}{
\pgfplotsset{                            %%% newly added for font family setting
      label style = {font=\footnotesize},
      tick label style  = {font=\footnotesize},
      every axis label/.append style = {font=\footnotesize}
}

\begin{tikzpicture}
%\tikzstyle{every node}=[font=\Huge]
   \begin{axis}[%
      view={30}{40},
      height=0.25\textwidth,
      width=0.32\textwidth,
      xlabel={$\Delta y$},
      ylabel={$\Delta x$},
      zlabel={$\bar{\mathcal{S}}(\mathcal{P}_{\Delta x, \Delta y}, \mathcal{Q})$},
      xmin = -120, xmax = 120,
      xtick={-100, -50, ..., 100},
      ymin = -120, ymax = 120,
      ytick={-100, -50, ..., 100},
	  ztick=\empty,
	  xtick align=inside,
	  ytick align=outside,
	  zlabel shift = 2ex,
  	  xlabel shift = -3ex,
  	  ylabel shift = -3ex,
      zlabel style={yshift=-2ex,xshift=2ex}      
   ]
\addplot3 [surf, mesh/rows=45, patch type=bilinear] table[x index=0, y index=1, z index=2] {pgf/data/38_4021_2d_bin.dat};
 \end{axis}
\end{tikzpicture}
}\\
\end{tabular}
\end{center}
\vspace{-2.5ex}
\caption{Alignment results for the example in Figure~\ref{fig:sketchmatch}. Similarity as a function of translation (in pixels): independent 1D projections (left), the full 2D translation (middle), and the 2D binarized polynomial (right). At zero translation the centers of the sketch and the image are aligned. The detection in magenta color (Figure~\ref{fig:sketchmatch}) is based on the similarity shown at the left, while the one in green is based on that shown in the middle.
\label{fig:align}
\vspace{-3ex}}
\end{figure*}
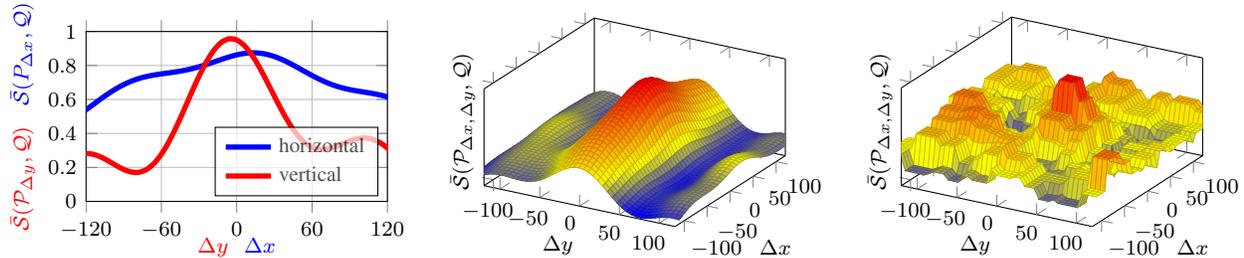
\paragraph{Two dimensional.}
Consider the full 2D translation $(\Dx, \Dy)$. Descriptor $\V(\Pc)$ encoding both spatial coordinates is used. The corresponding second order trigonometric polynomial~\cite{TBFJ15} of scores $\sim(\Pc_{\!\!\Dx, \Dy}, Q)$ is constructed similarly to the first order one. The details are omitted for the sake of brevity.
It allows for an efficient evaluation of similarity for multiple 2D translations in a sliding window manner. 
The cost to compute one of its coefficients is $4 D_\o$.
There are $N_2=4(|\Omega_x|-1)^2 +4(|\Omega_x|-1) + 1$ non-zero coefficients in total. 
The similarity computation for a single 2D translation has cost equal to $N_2$ scalar multiplications. Translation invariant similarity is given by $\sim_{xy}(\Pc_{\Dx, \Dy}, \Qc) = \max_{(\Dx, \Dy)} \sim(\Pc_{\Dx,\Dy}, \Qc)$, and the transformation aligning the two shapes is given by $(\hat{x}_2,\hat{y}_2) = \arg\max_{(\Dx, \Dy)} \sim(\Pc_{\Dx,\Dy}, \Qc)$.

In Figures~\ref{fig:sketchmatch} and~\ref{fig:align} we present an alignment example between a sketch and a real image. Similarity is computed based on the horizontal and vertical projections, while also for the 2D case. Maximum similarity is met at translations that align the two silhouettes. 

\subsection{Efficient retrieval and query expansion}
Herein, we propose three methods how to avoid exhaustive evaluation of $\sim(\Pc_{\Dx,\Dy}, \Qc)$.
First method efficiently selects a shortlist of images on which the score $\bar{\sim}_{xy}$ is computed. The other two methods are designed to limit the number of possible translations over which 
$\sim(\Pc_{\Dx,\Dy}, \Qc)$
is evaluated to obtain a good approximation of $\bar{\sim}_{xy}$. 

\paragraph{Shortlist by projections.}
The similarities $\bar{\sim}_x$ and $\bar{\sim}_y$ computed over the projections (\ref{equ:tripol1}) provide an estimate of the $\bar{\sim}_{xy}$. We propose to use this estimate for initial ranking and to compute the slow similarity $\bar{\sim}_{xy}$ only on a shortlist of top $S$ images. Experiments show that initial ranking by $\bar{\sim}_x +\bar{\sim}_y$ outperforms ranking that uses only one projection. To further speed-up the evaluation for large-scale collections, we propose {\em discriminative projection first} approach. In this method, one projection is computed over the whole dataset, creating a pre-shortlist of $3 S$ images with the highest score. The second projection is only evaluated on this pre-shortlist. Now, the shortlist based on the value of $\bar{\sim}_x +\bar{\sim}_y$ is a sub-set of the pre-shortlist. The first projection used is query dependent, the one with higher variance in the relevant coordinate in the sketch query is used. Discriminative projection first is denoted as \dpf.

\paragraph{Re-ranking by local refinement.}
The 1D alignment provides, besides the scores, the scale and 1D translations $(\hat{x}_1, \hat{y}_1)$ maximizing the 1D projection scores, which often is a rough approximation of the full 2D alignment (see Figure~\ref{fig:sketchmatch} and~\ref{fig:align}). In this approach, the full similarity 
$\sim(\Pc_{\Dx,\Dy}, \Qc)$ 
is only evaluated evaluated for a small neighborhood of $(\hat{x}_1, \hat{y}_1)$ on a fixed 2D grid. Sketch similarity computed by this method is denoted by  $\sim_{x/y}$.

\paragraph{Re-ranking by binary polynomial.} 
We efficiently approximate the second order polynomial 
by a corresponding one that has binary coefficients and variables (\ie $\cos(\omega_x \Dx)\cos(\omega_y \Dy)$ is binarized). We simply binarize both by a sign function. The similarity approximation for 2D translation is given by dot product between binary vectors which is faster to compute. 
Translation maximizing the binary approximation is found, and 
$\sim(\Pc_{\Dx,\Dy}, \Qc)$ 
is only computed on a small neighborhood, as in the local refinement. Figure~\ref{fig:align} shows an example where the position of the maximum on the 2D map of similarities for the binarized case remains close to that of the real valued one. Experiments show that the binary polynomials provide very good estimate of the translation. We denote this method by  $\sim_{xy\star}$. 

\paragraph{Query expansion.} Query Expansion (QE) is a standard approach to improve retrieval results by a new query that exploits the top-ranked results~\cite{CPSIZ07,DGBQG11,JB09}. 
Unlike the original query, the QE is performed on image descriptors, the sketch descriptors are only used for localization.
A global CNN image descriptor is used for QE, in particular off-the-shelf CroW~\cite{KMO16} with VGG16 network~\cite{SZ14}. The 512D image descriptor extracted per database image is compressed using product quantization~\cite{JDS11} into 64 bytes. A basic version of an Average Query Expansion (AQE)~\cite{CPSIZ07} is used. CroW descriptors of the top results are averaged and a query is issued. 

\vspace{-5pt}
\section{Experiments}
\label{sec:exp}
\vspace{-5pt}
We briefly summarize the design choices of the indexing and search procedure of our sketch-based retrieval. Then, we evaluate our method and compare to the state of the art. 

\label{sec:retrieval}
\paragraph{Indexing {\rm (offline stage).}} All database images are down-sampled to have the longer side equal to 400 pixels. 
The edges are detected by off-the-shelf detector of Doll{\'a}r and Zitnick~\cite{DZ13}. 
The output edge strength is used as $\pw$, while all edges with strengths lower than 0.2 are completely discarded. 
A single sketch descriptor per database image is computed with AFM (\ref{equ:map_asym2}). 
Three kernels are used to search at three scales.
Finally, the corresponding \l2 norms for normalizing similarity (\ref{equ:cossim}) are computed and stored.

\paragraph{Query {\rm (online stage).}} The sketch query is cropped with a tight bounding box and resized similarly to database images.
Two additional  scales are given by down-sampling to $80\%$ and $60\%$. 
Different query scales need to be matched with different kernels; smaller scale is matched with narrower kernel. 
The kernels shown in Figure~\ref{fig:jointapprox} are used accordingly. 
The orientation kernel has $\sigma_\o=0.8$. 
One query descriptor per kernel is constructed (\ref{equ:map_asym1}). 
Additionally, each query is also horizontally mirrored.

The translations to be evaluated are fixed in a uniform way. 
Maximum translation is set to 80 pixels towards both directions and the step is 20. 
These are used for the maximum query size, while for different scales the maximum translation (step) is increased (decreased) linearly according to the relative query scale. 
That means, the localization is finer for smaller scales. Similarity is computed per scale independently and maximum similarity is kept.

\paragraph{The descriptor dimensionality} is given by the number of frequencies $|\Omega_x|$ and $|\Omega_\phi|$.  
For instance, a compact setting of $|\Omega_x| = 5$ and $|\Omega_\phi| = 2$ lead to a 243D descriptor, while a high-performance settings of
$|\Omega_x| = 6$ and $|\Omega_\phi| = 3$ lead to a 605D descriptor.
In all cases, 9 additional scalars per image are stored (normalization of the 2D descriptor, normalizations of the 1D projections, all for 3 different scales).

\paragraph{Method identification.} 
The following notation is used to identify the method, \emph{ranking method} $\rightarrow$ \emph{re-ranking method} (\emph{number of re-ranked images}). 
Usage of average query expansion using $n$ top images is denoted by QE$n$.

\subsection{Datasets and evaluation protocol}
Constructing large scale ground-truth for sketch-based retrieval systems is not as easy as for traditional retrieval. One reason is the inherent abstraction of sketches. Moreover, positive images should not only comprised images of the same object/category, but also images depicting shapes similar to that of the query. Ground-truth at large scale should be on per query basis and this is not easy to achieve.

We initially evaluate our method on two image collections that are accompanied with ground-truth. 
These are the ETHZ extended shape dataset~\cite{SS08} and the Flickr15k dataset~\cite{RC13}. They consist of 285 images with 7 queries and 15K images with 330 queries (30 categories), respectively. 
 
We further perform experiments on the large-scale dataset by Parui and Mittal~\cite{PM14} comprised 1.2M images and 175 queries, which has no available annotation. External annotators have manually evaluated the top results. 
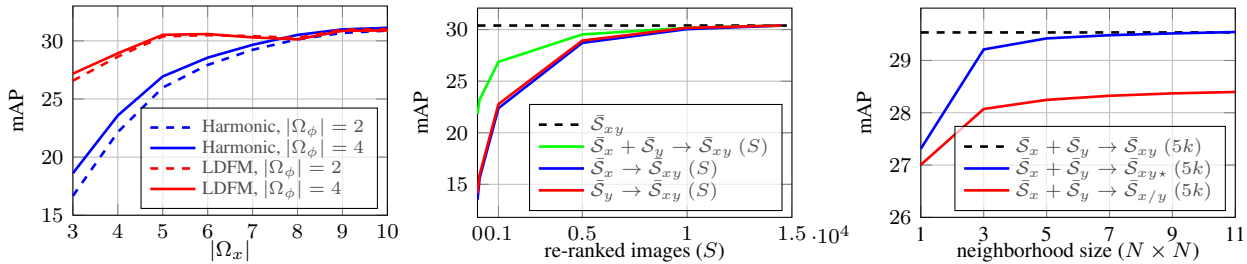
\begin{figure*}[t]
\vspace{-5ex}
\begin{center}
\extfig{map_vs_nf}{
\pgfplotsset{                            %%% newly added for font family setting
      label style = {font=\footnotesize},
      tick label style  = {font=\footnotesize},
      every axis label/.append style = {font=\footnotesize}
}
\begin{tikzpicture}
% \tikzstyle{every node}=[font=\huge]
   \begin{axis}[%
      height=0.25\textwidth,
      width=0.33\textwidth,
      xlabel={$|\Omega_x|$},
      ylabel={mAP},
      legend pos=south east,
      legend cell align=left,
      legend style={font=\scriptsize, fill opacity=0.7, row sep=-3pt},
      grid=both,
      ymin = 15, ymax = 33,
	   ytick = {15, 20, ..., 35},
      xmin = 3, xmax = 10,
      xtick={3, 4, ..., 10},
      ytick={15,20,25,30},
      xlabel style={yshift=2ex},      
      ylabel style={yshift=-3ex}
   ]
\addplot[color=blue,dashed, line width=1pt] table[x expr={1+\thisrowno{0}},y expr={100*\thisrowno{2}}]{pgf/data/fourier_map_vs_n.dat};
\leg{Harmonic, $|\Omega_\phi|=2$}
\addplot[color=blue, line width=1pt] table[x expr={1+\thisrowno{0}},y expr={100*\thisrowno{6}}]{pgf/data/fourier_map_vs_n.dat};
\leg{Harmonic, $|\Omega_\phi|=4$}
\addplot[color=red,dashed, line width=1pt] table[x expr={1+\thisrowno{0}},y expr={100*\thisrowno{2}}]{pgf/data/ldfm_map_vs_n.dat};
\leg{LDFM, $|\Omega_\phi|=2$}
\addplot[color=red, line width=1pt] table[x expr={1+\thisrowno{0}},y expr={100*\thisrowno{6}}]{pgf/data/ldfm_map_vs_n.dat};
\leg{LDFM, $|\Omega_\phi|=4$}
\end{axis}
\end{tikzpicture}
}\hspace{-3ex}
\extfig{map_align2D_vs_align1D}{
\pgfplotsset{                            %%% newly added for font family setting
      label style = {font=\footnotesize},
      tick label style  = {font=\footnotesize},
      every axis label/.append style = {font=\footnotesize}
}
\begin{tikzpicture}
   \begin{axis}[
      height=0.25\textwidth,
      width=0.33\textwidth,
      xlabel={re-ranked images ($S$)},
      ylabel={mAP},
      legend pos=south east,
      legend cell align=left,
      legend style={font=\scriptsize, fill opacity=0.7, row sep=-3pt},
      grid=both,
      xmin = 0, xmax = 15000,
      xtick= {0, 1000, 5000, 10000, 15000},
      xlabel style={yshift=2ex},      
      ylabel style={yshift=-3ex},      
      every x tick scale label/.style={at={(1,0)},xshift=7pt,yshift=-10pt,anchor=south west,inner sep=0pt}
   ]
\addplot[color=black,dashed, mark = none, samples = 2, line width=1pt] coordinates{(1,30.396) (15000, 30.396
)};\leg{$\bar{\mathcal{S}}_{xy}$}
\addplot[color=green, line width=1pt] table[x index=0, y expr={100*\thisrowno{1}}]{pgf/data/map_align1D.dat};
\leg{$\bar{\mathcal{S}}_x+\bar{\mathcal{S}}_y \rightarrow \bar{\mathcal{S}}_{xy}~(S)$}
\addplot[color=blue, line width=1pt] table[x index=0,y expr={100*\thisrowno{2}}]{pgf/data/map_align1D.dat};
\leg{$\bar{\mathcal{S}}_x \rightarrow \bar{\mathcal{S}}_{xy}~(S)$}
\addplot[color=red, line width=1pt] table[x index=0,y expr={100*\thisrowno{3}}]{pgf/data/map_align1D.dat};
\leg{$\bar{\mathcal{S}}_y \rightarrow \bar{\mathcal{S}}_{xy}~(S)$}
\end{axis}
\end{tikzpicture}
}\hspace{-3ex}
\extfig{map_rerank_approx}{
\pgfplotsset{                            %%% newly added for font family setting
      label style = {font=\footnotesize},
      tick label style  = {font=\footnotesize},
      every axis label/.append style = {font=\footnotesize}
}
\begin{tikzpicture}
   \begin{axis}[%
      height=0.25\textwidth,
      width=0.33\textwidth,
      xlabel={neighborhood size ($N\times N$)},
      ylabel={mAP},
      legend pos=south east,
      legend cell align=left,
      legend style={font=\scriptsize, fill opacity=0.7, row sep=-3pt},
      grid=both,
      xmin = 0, xmax = 5,
      xtick= {0, 1, ..., 5},
      ymin = 26, ymax = 30,
      ytick= {26,27,28,29},
      xticklabels={1,3,5,7,9,11},
      xlabel style={yshift=2ex},      
      ylabel style={yshift=-3ex}      
   ]
\addplot[color=black,dashed, mark = none, samples = 2, line width=1pt] coordinates{(0,29.535) (5, 29.535)};\leg{$\bar{\mathcal{S}}_{x}+\bar{\mathcal{S}}_{y} \rightarrow \bar{\mathcal{S}}_{xy}~(5k)$}
\addplot[color=blue, line width=1pt] table[x index=0,y expr={100*\thisrowno{1}}]{pgf/data/map_rerank_approx.dat};\leg{$\bar{\mathcal{S}}_{x}+\bar{\mathcal{S}}_{y} \rightarrow \bar{\mathcal{S}}_{xy\star}~(5k)$}
\addplot[color=red, line width=1pt] table[x index=0,y expr={100*\thisrowno{2}}]{pgf/data/map_rerank_approx.dat};\leg{$\bar{\mathcal{S}}_{x}+\bar{\mathcal{S}}_{y} \rightarrow \bar{\mathcal{S}}_{x/y}~(5k)$}
\end{axis}
\end{tikzpicture}
}
\end{center}
\vspace{-3.5ex}
\caption{Performance comparison by measuring mean Average Precision (mAP)  on the Flickr15k dataset. \textbf{Left}: Performance for increasing number of frequencies. Comparison between the Fourier-based approach~\cite{VZ12} that uses harmonic frequencies and our joint optimization of the 3 kernel functions. Ranking is performed with $\bar{\sim}_{xy}$.
\textbf{Middle}: Comparison between the proposed methods for ranking the whole dataset. Re-ranking is additionally performed with $\bar{\sim}_{xy}$ in all cases. We show mAP versus the number of re-ranked images. $S=0$ signifies no re-ranking. 
\textbf{Right}: Performance of approximate re-ranking methods for increasing size of local refinement neighborhood. We show mAP versus the neighborhood size, while re-ranking 5k images. 
\label{fig:exp_triplet}
\vspace{-8pt}}
\end{figure*}
\begin{figure*}[t]
\setlength{\fboxsep}{0pt}%
\setlength{\fboxrule}{2pt}%
\centering

\pgfplotstableread{fig/ranks_annot_h.txt}\table
\pgfplotstablegetrowsof{\table}
\pgfmathsetmacro{\M}{\pgfplotsretval-1}
\pgfplotstablegetcolsof{\table}
\pgfmathsetmacro{\N}{\pgfplotsretval-1}

\begin{tabular}{@{\sssp}c@{\sssp}c@{\sssp}}
\def\query{10}
\includegraphics[height=8mm]{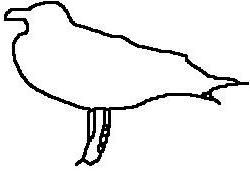}\vspace{0.01cm}
\foreach \c in {1,...,5}  {
\pgfplotstablegetelem{9}{[index]\c}\of\table
\ifnum\pgfplotsretval=3
\fcolorbox{yellow}{black}{\includegraphics[height=8mm]{fig/select/\query_\c_align.png}\vspace{0.01cm}}
\else
\ifnum\pgfplotsretval=2
\fcolorbox{green}{black}{\includegraphics[height=8mm]{fig/select/\query_\c_align.png}\vspace{0.01cm}}
\else
\fcolorbox{red}{black}{\includegraphics[height=8mm]{fig/select/\query_\c_align.png}\vspace{0.01cm}}
\fi\fi
}
&
\def\query{87}
\includegraphics[height=8mm]{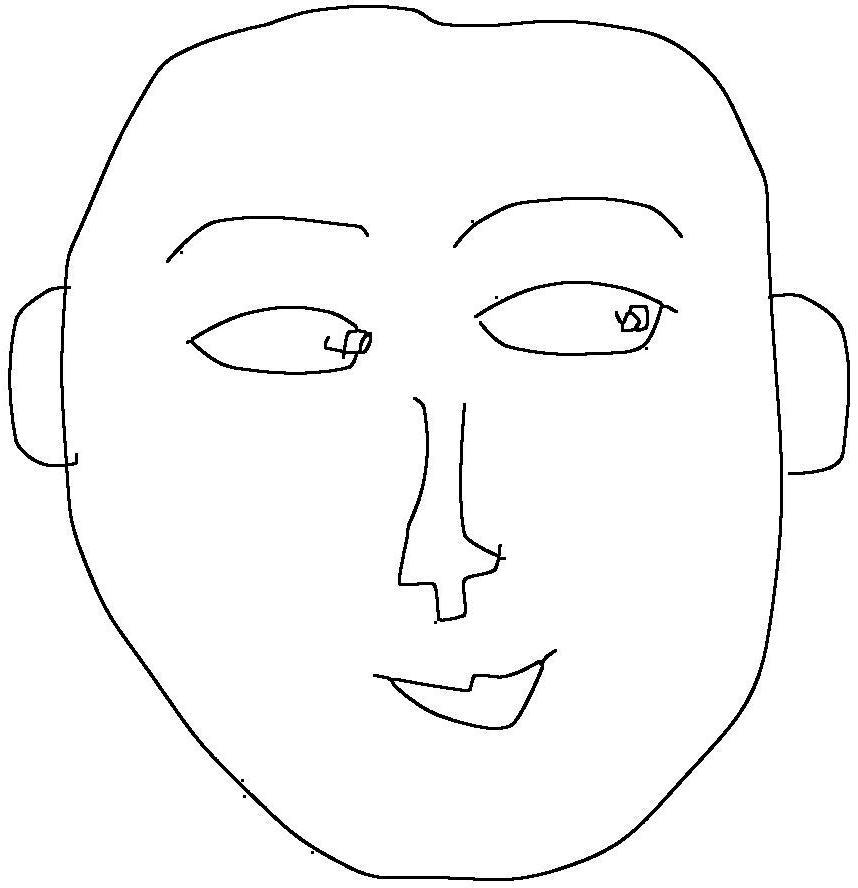}\vspace{0.01cm}
\foreach \c in {1,...,6}  {
\pgfplotstablegetelem{86}{[index]\c}\of\table
\ifnum\pgfplotsretval=3
\fcolorbox{yellow}{black}{\includegraphics[height=8mm]{fig/select/\query_\c_align.png}\vspace{0.01cm}}
\else
\ifnum\pgfplotsretval=2
\fcolorbox{green}{black}{\includegraphics[height=8mm]{fig/select/\query_\c_align.png}\vspace{0.01cm}}
\else
\fcolorbox{red}{black}{\includegraphics[height=8mm]{fig/select/\query_\c_align.png}\vspace{0.01cm}}
\fi\fi
}\\
\def\query{45}
\includegraphics[height=8mm]{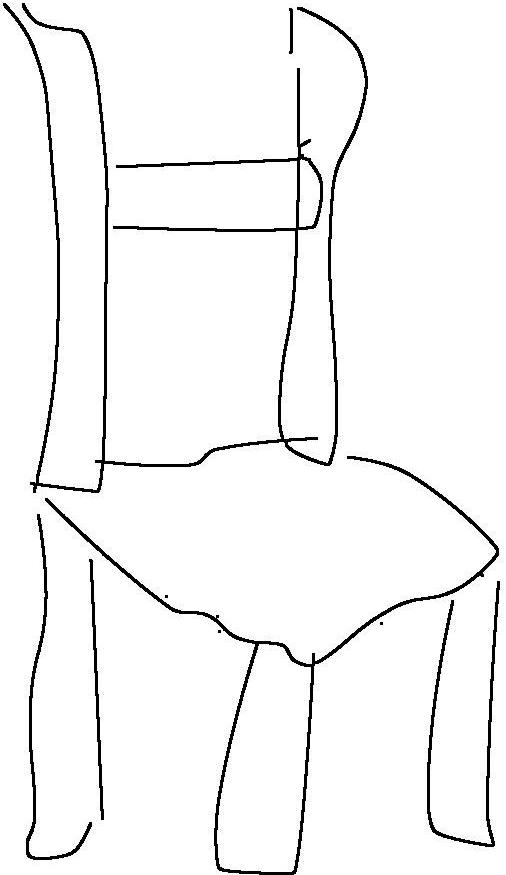}\vspace{0.01cm}
\foreach \c in {1,...,7}  {
\pgfplotstablegetelem{44}{[index]\c}\of\table
\ifnum\pgfplotsretval=3
\fcolorbox{yellow}{black}{\includegraphics[height=8mm]{fig/select/\query_\c_align.png}\vspace{0.01cm}}
\else
\ifnum\pgfplotsretval=2
\fcolorbox{green}{black}{\includegraphics[height=8mm]{fig/select/\query_\c_align.png}\vspace{0.01cm}}
\else
\fcolorbox{red}{black}{\includegraphics[height=8mm]{fig/select/\query_\c_align.png}\vspace{0.01cm}}
\fi\fi
}
&
\def\query{95}
\includegraphics[height=8mm]{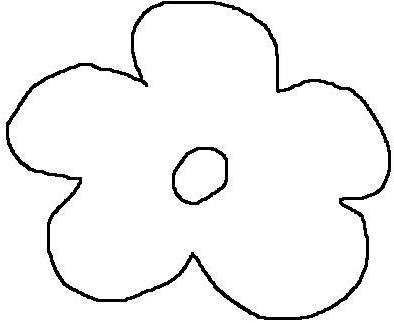}\vspace{0.01cm}
\foreach \c in {1,...,5}  {
\pgfplotstablegetelem{94}{[index]\c}\of\table
\ifnum\pgfplotsretval=3
\fcolorbox{yellow}{black}{\includegraphics[height=8mm]{fig/select/\query_\c_align.png}\vspace{0.01cm}}
\else
\ifnum\pgfplotsretval=2
\fcolorbox{green}{black}{\includegraphics[height=8mm]{fig/select/\query_\c_align.png}\vspace{0.01cm}}
\else
\fcolorbox{red}{black}{\includegraphics[height=8mm]{fig/select/\query_\c_align.png}\vspace{0.01cm}}
\fi\fi
}\\
\def\query{63}
\includegraphics[height=8mm]{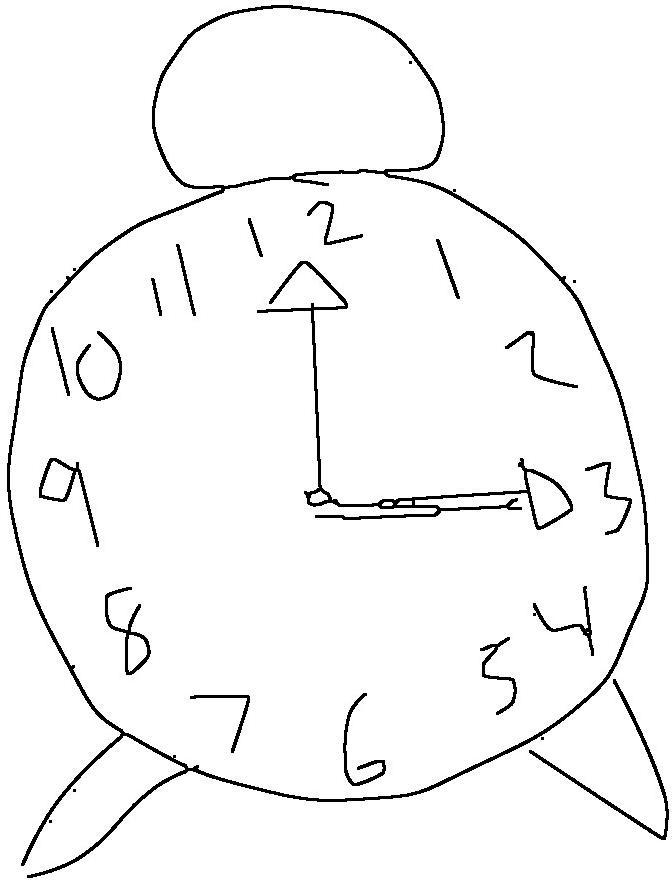}\vspace{0.01cm}
\foreach \c in {1,...,7}  {
\pgfplotstablegetelem{62}{[index]\c}\of\table
\ifnum\pgfplotsretval=3
\fcolorbox{yellow}{black}{\includegraphics[height=8mm]{fig/select/\query_\c_align.png}\vspace{0.01cm}}
\else
\ifnum\pgfplotsretval=2
\fcolorbox{green}{black}{\includegraphics[height=8mm]{fig/select/\query_\c_align.png}\vspace{0.01cm}}
\else
\fcolorbox{red}{black}{\includegraphics[height=8mm]{fig/select/\query_\c_align.png}\vspace{0.01cm}}
\fi\fi
}
&
\def\query{103}
\includegraphics[height=8mm]{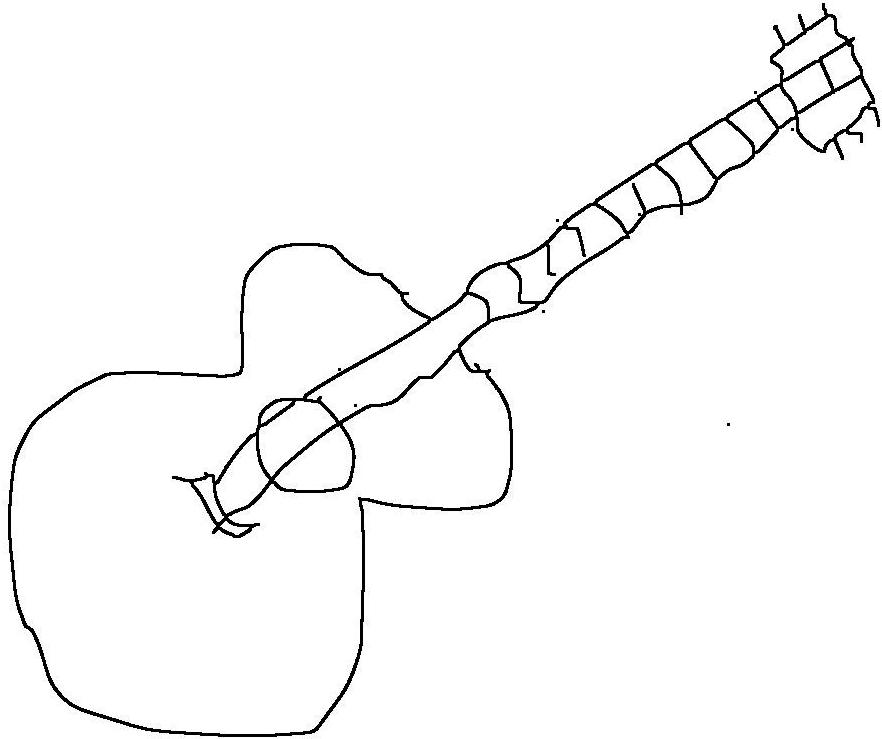}\vspace{0.01cm}
\foreach \c in {1,...,5}  {
\pgfplotstablegetelem{102}{[index]\c}\of\table
\ifnum\pgfplotsretval=1
\fcolorbox{red}{black}{\includegraphics[height=8mm]{fig/select/\query_\c_align.png}\vspace{0.01cm}}
\else
\ifnum\pgfplotsretval=2
\fcolorbox{green}{black}{\includegraphics[height=8mm]{fig/select/\query_\c_align.png}\vspace{0.01cm}}
\else
\fcolorbox{yellow}{black}{\includegraphics[height=8mm]{fig/select/\query_\c_align.png}\vspace{0.01cm}}
\fi\fi
}\\
\end{tabular}

\caption{Examples of top-ranked retrieval images on the 1.2M dataset using our method. Localization of the sketch is shown in green color. Image borders denote positive (green), negative (red) and similar (yellow) image.\label{fig:example_large}
 \vspace{-12pt}}
\end{figure*}
\begin{table}[t]
\vspace{-6pt}
\small
\centering
\setlength\extrarowheight{-1pt}
\begin{tabular}{|@{\msp}l@{\msp}|@{\msp}c@{\msp}|@{\msp}l@{\msp}|@{\msp}c@{\msp}|}
\hline
Method	 							&   	 	P@20   	&  Method 									&     		P@20  	\\ \hline \hline
EI~\cite{CWZZ11}					&    	 	27.9 		&  $\bar{\sim}_{xy}~(5,2)$				& 	   	  	57.9 		\\
Riemenschneider~\cite{RDB11}	& 		   58.0		&  $\bar{\sim}_{xy}~(6,3)$				& 	 			61.4	 	\\
SYM-FISH~\cite{CZL+13} 		 	& 	  		34.0 		&  $\bar{\sim}_{xy}~(5,2)$ + QE3   & 	   	  	77.9 		\\
CS+GC~\cite{PM14}					& 	  		49.3		&  $\bar{\sim}_{xy}~(6,3)$ + QE3	&   \textbf{79.3}    \\
 \hline 
\end{tabular}
\caption{Performance comparison on the ETHZ extended shape dataset. Average precision at top 20 results is reported. We have not performed query mirroring for these results. The number of frequencies $(|\Omega_x|,|\Omega_\o|)$ used is reported next to our methods.\label{tab:ethz}}
\vspace{-10pt}
\end{table}
\begin{table}[t]
\small
\centering
\setlength\extrarowheight{1pt}
\begin{tabular}{|@{\msp}l@{\msp}|@{\msp}c@{\msp}|@{\msp}l@{\msp}|@{\msp}c@{\msp}|}
\hline
Method 										&    		mAP    		& 	Method 																				&    mAP    			\\ \hline \hline
GF-HOG~\cite{RC13} 						&   		12.2    		& 	$\bar{\sim}_x + \bar{\sim}_y \rightarrow \bar{\sim}_{xy}$ (1k) 	& 	26.7    			\\
SHELO~\cite{Saaverda14}  				& 			12.3    		& 	$\bar{\sim}_x + \bar{\sim}_y \rightarrow \bar{\sim}_{xy}$ (5k)	   & 	29.2    			\\
LKS~\cite{SBO15}							& 			24.5	  		& 	$\bar{\sim}_{xy}$ 																& 	30.4			 	\\
GF-HOG~\cite{BC15}	         		&        18.2        &  $\bar{\sim}_{xy}$ + QE3														&  \textbf{57.9}	\\
\hline 
\end{tabular}
\caption{Performance comparison via mean Average Precision on the Flickr15k dataset.\label{tab:flickr15k}}
\vspace{-10pt}
\end{table}
\subsection{Evaluation and comparisons}
\paragraph{Performance versus dimensionality.}
We construct the proposed sketch descriptor using our LDFM-based multiple kernel approximation and using the Fourier-based one. We compare performance for varying number of frequencies and present results in Figure~\ref{fig:exp_triplet} (left). The two methods have roughly the same performance for large number of frequencies where the kernel approximation is relatively good for both cases. The Fourier-based method significantly harms the performance for low number of frequencies due to its bad approximation. The orientation kernel is well approximated with few frequencies due to its wide shape (larger $\sigma$). We finally set $|\Omega_x|=5$ and $|\Omega_\o|=2$ for the rest of our experiments, except if otherwise stated. Sketch descriptor $\V(\Pc)$ has 243 dimensions, while $\V_x(\Pc)$ only 27.

\paragraph{Ranking method.}
We compare ranking of the database with $\bar{\sim}_{xy}$ and the projection-based approaches $\bar{\sim}_{x}$ and $\bar{\sim}_{y}$. In the latter case, only the top-ranked images are re-ranked by $\bar{\sim}_{xy}$ to evaluate the performance loss. Results are shown in Figure~\ref{fig:exp_triplet} (middle). Ranking with sum of $\bar{\sim}_{y}$ and $\bar{\sim}_{x}$ appears  significantly better than their individual use, while re-ranking one third of the database already recovers the performance loss.
Speeding-up the ranking by \dpf, while in the end we re-rank 1k images, achieves mAP equal to 26.8. The drop is insignificant compared to the 26.9 in Figure~\ref{fig:exp_triplet} when re-ranking 1k images. Always ranking first with $x$ or $y$ projection, instead of our query dependent approach, gives 25.8 and 26.0 respectively. 

\paragraph{Approximations.}
We perform re-ranking based on $\bar{\sim}_{xy}$ and its two approximations. We use approximation $\bar{\sim}_{xy\star}$ to efficiently search over all translations and scales, while we finally refine the translation of maximum similarity. On the other hand, $\bar{\sim}_{x/y}$ is used to refine $(\hat{x}_1, \hat{y}_1)$ and acts only on the best scale found by the ranking method. In some cases the ranking method misses the  correct scale and this is the main reason for the performance difference between the two. Results are shown in Figure~\ref{fig:exp_triplet} (right).

\paragraph{Comparisons to other methods.}
Comparison of our method to other methods is reported in Table~\ref{tab:ethz} for the ETHZ extended shape dataset and in Table~\ref{tab:flickr15k} for the Flickr15k dataset.
The scores achieved without the QE are the highest reported on both benchmarks. The QE gives additional significant boost in the performance. On Flickr15k we remarkably outperform the previous state of the art by 24 points of mAP. 
\begin{table}[t]
\vspace{-7pt}
\newcommand{\spc}{\hspace{.8em}\mbox{}}
\footnotesize
\setlength\extrarowheight{1pt}
\definecolor{Gray}{gray}{0.85}
\centering
\begin{tabular}{|@{\sssp}l@{\sssp}|@{\sssp}r@{\sssp}|@{\sssp}r@{\sssp}|@{\sssp}r@{\ssp}|@{\hspace{-.5pt}}r@{\hspace{-.5pt}}|@{\sssp}r@{\sssp}|@{\sssp}r@{\sssp}|@{\sssp}r@{\sssp}|@{\sssp}r@{\sssp}|}
\hline 
Method  									  																			& Dim				& Time &  DB  	&& P@5  &  @10    & @25    & @50  \\ \hline \hline
$\bar{\sim}_{xy}$ (1.2M){\cancel{AFM}}~\cite{TBFJ15}  												& (8,3)   	   & 55.4 &  15.3 && 43.2 &  40.9   & 37.2   & 33.8 \\ 
$\bar{\sim}_{xy}$ (1.2M){\cancel{AFM}}~\cite{TBFJ15}  												& (5,2)			& 20.2 &  3.3  && 25.8 &  24.7   & 22.5   & 20.2 \\ \hline
$\bar{\sim}_{xy}$ (1.2M)																						& (8,3) 			& 55.4 &  5.1	&& 50.1 &  46.7   & 42.0   & 37.2 \\
$\bar{\sim}_{xy}$ (1.2M)																						& (5,2)			& 20.2 &  1.1  && 45.8 &  44.1   & 38.5   & 35.4 \\ \hline
$\bar{\sim}_x\! +\! \bar{\sim}_y ${\scriptsize$\rightarrow$}$ \bar{\sim}_{xy\star}$ (50k)	& (6,3)			& 3.5  &  2.8  && 49.7 &  47.4   & 41.3   & 36.8 \\
\dpf {\scriptsize$\rightarrow$}$ \bar{\sim}_{xy\star}$ (50k)										& (6,3)   		& 2.5  &  2.8  && 49.6 &  47.3   & 41.0   & 36.6 \\
\dpf {\scriptsize$\rightarrow$}$ \bar{\sim}_{xy\star}$ (50k)$^\dagger$							& (6,3)	      & 2.5  &  0.7	&& 50.3 &  47.3	& 41.5   & 36.7 \\		
$\bar{\sim}_x\! +\! \bar{\sim}_y ${\scriptsize$\rightarrow$}$ \bar{\sim}_{xy\star}$ (50k)	& (5,2)	  		& 2.5  &  1.1  && 45.8 &  44.2   & 38.4   & 35.3 \\
\dpf {\scriptsize$\rightarrow$}$ \bar{\sim}_{xy\star}$ (50k)										& (5,2)	  		& 1.7  &  1.1  && 45.7 &  44.2   & 38.3   & 35.1 \\
\dpf {\scriptsize$\rightarrow$}$ \bar{\sim}_{xy\star}$ (50k)$^\dagger$							& (5,2)        & 1.7  &  0.3	&& 45.6 &  43.5   & 38.0   & 35.0 \\		
 \hline %\hline 
\dpf {\scriptsize$\rightarrow$}$ \bar{\sim}_{xy\star}$ (50k)$^\dagger$+QE3						& (6,3)		   & 2.7  &  0.8  && 55.2 &  57.4   & 57.4   & 57.5  \\ 
\dpf {\scriptsize$\rightarrow$}$ \bar{\sim}_{xy\star}$ (50k)$^\dagger$+QE10					& (6,3)        & 2.7  &  0.8  && 63.0 &  63.4   & 64.8   & 65.2  \\ 
\dpf {\scriptsize$\rightarrow$}$ \bar{\sim}_{xy\star}$ (50k)$^\dagger$+QE3						& (5,2)		   & 1.9  &  0.4  && 50.9 &  52.2   & 52.5   & 52.4  \\
\dpf {\scriptsize$\rightarrow$}$ \bar{\sim}_{xy\star}$ (50k)$^\dagger$+QE10					& (5,2)        & 1.9  &  0.4  && 56.4 &  56.8   & 57.3   & 57.8  \\  \hline % 1 PQ query 
\end{tabular}	
% \vspace{1ex}
\xcaption[0]{Performance, query time (in seconds) and database (DB) memory (in GB) requirements comparison on the 1.2M image dataset~\cite{PM14}. We report precision at $n$ top ranked images (P@$n$). The number of frequencies ($|\Omega_x|,|\Omega_\o|$) is reported, which defines the final dimensionality (Dim = 1125, 605 or 243).
\cancel{AFM}: Asymmetric feature maps are not used. $\dagger$: Vector components uniformly quantized into 1 byte.
\label{tab:largescale}
\vspace{-15pt}}
\end{table}

\paragraph{Large scale evaluation.}
We evaluate our method at large scale with the 1.2M dataset~\cite{PM14}. For each query, only top-ranked images are annotated as either negative, positive or similar. Images marked as similar are images of similar shape but different category than the query.
Retrieval examples are shown in Figure~\ref{fig:example_large} and performance comparison
is presented in Table~\ref{tab:largescale}. 
We measure precision at top-ranked images per query and report average precision on top ranked images over all queries.

We additionally evaluate performance when applying the trigonometric polynomial of Tolias \etal~\cite{TBFJ15} to rank all database images. 
The proposed method by construction requires less memory and is significantly faster. It is also shown to perform better. 
The memory footprint is significantly decreased due to the asymmetry of our representation and due to good performance achieved with few frequencies.
Encoding each vector component with 1 byte instead of single precision does not harm the performance.

Note that the discriminative-projection-first method only slightly decreases the performance, while it decreases the initial ranking time by $40\%$.
Moreover, re-ranking only top 50k images performs with insignificant losses compared to ranking all images with the 2D polynomial.  
Finally, query expansion significantly improves the results. CNN descriptors are encoded with product quantization~\cite{JDS11}\footnote{After evaluation we discovered that the dataset contains a small amount of training ImageNet images, which can potentially affect with QE by CNN descriptors. Preliminary tests show that it affects insignificantly.}.

\paragraph{Query timings.} The execution time was measured on the 1.2M image dataset using a single threaded MATLAB/Mex implementation on a 3.5GHz desktop machine. The results are summarized in Table~\ref{tab:largescale}. For $|\Omega_x| = 5$ and $|\Omega_\phi| = 2$, a query takes on average $1.81$s for the initial ranking with $\bar{\sim}_x + \bar{\sim}_y$ and $0.72$s for the top 50k re-ranking with $\bar{\sim}_{xy\star}$ (with 3x3 neighborhood), giving a total time of $2.5$s. Using \dpf, and computing the second projection only on $3 \cdot 50$k top-ranked images, ranking time drops to $1.05$s. The values are independent of query complexity. The re-ranking using binary $\bar{\sim}_{xy\star}$ is $17\%$ faster compared to full $\bar{\sim}_{xy}$. 
Applying the trigonometric polynomial scoring for ranking all images with the method of Tolias \etal~\cite{TBFJ15} takes $20$s, one order of magnitude slower than ours, for a low performance setup, while $55$s with higher dimensionality and better performance which is still lower than ours. 

The performance comparison to the work of Parui and Mittal~\cite{PM14} is not possible on the 1.2M dataset, as they use their own category-level ground truth, which is not publicly available. The comparison in terms of memory footprint (6.5GB is reported\cite{PM14} ) and execution time (1-5 sec per query is reported\cite{PM14}) is favorable for the proposed method.

\vspace{-5pt}
\section{Conclusions}
\vspace{-5pt}
We have introduced a novel concept of asymmetric (explicit) feature maps. AFM allow to evaluate multiple kernels between a query and database entries with no additional memory requirements. The feature maps are optimally constructed by a joint kernel approximation, which turns out to be crucial for the accuracy. We have introduced a method of efficient approximation of scoring by trigonometric polynomials through 1D projections, which are a special case of asymmetric feature maps. 

We have demonstrated the benefits of AFM on sketch-based image retrieval with short codes. We achieve state-of-the-art performance on a number of standard benchmarks. Compared with previous approaches using trigonometric polynomials~\cite{TBFJ15}, the proposed method achieves an order of magnitude speed-up, multiple-fold reduction in data storage, while improving the retrieval accuracy at the same time.
The performance is further boosted by image-based average
query expansion combined with AFM for object outline localization.

{\small
\bibliographystyle{ieee}
\bibliography{egbib}
}
%\flushend
\end{document}